\documentclass{bmcart}

\usepackage[utf8]{inputenc} %unicode support
\def\includegraphics{}
\usepackage{graphicx}

\usepackage{hyperref}
\usepackage{amsmath} 
\usepackage{booktabs}
\usepackage{algorithm}
\usepackage{algpseudocode}
\usepackage{amsmath}
\usepackage{soul} 
\usepackage{lmodern}

\begin{document}

%%% Start of article front matter
\begin{frontmatter}

\begin{fmbox}
\dochead{Methodology}


\title{FrogDeepSDM: Improving Frog Counting and Occurrence Prediction Using Multimodal Data and Pseudo-Absence Imputation}

%%%%%%%%%%%%%%%%%%%%%%%%%%%%%%%%%%%%%%%%%%%%%%
%%                                          %%
%% Enter the authors here                   %%
%%                                          %%
%% Specify information, if available,       %%
%% in the form:                             %%
%%   <key>={<id1>,<id2>}                    %%
%%   <key>=                                 %%
%% Comment or delete the keys which are     %%
%% not used. Repeat \author command as much %%
%% as required.                             %%
%%                                          %%
%%%%%%%%%%%%%%%%%%%%%%%%%%%%%%%%%%%%%%%%%%%%%%

\author[
   addressref={aff2, aff3},                   % id's of addresses, e.g. {aff1,aff2}
   corref={aff2},                       % id of corresponding address, if any
   %noteref={n1},                        % id's of article notes, if any
   email={c.padubidri@cyens.org.cy}   % email address
]{\inits{CP}\fnm{Chirag} \snm{Padubidri}}
\author[
   addressref={aff1},
   % email={a.kamilaris@cyens.org.cy}
]{\inits{AK}\fnm{Pranesh} \snm{Velmurugan}}
\author[
   addressref={aff2, aff3},
   email={a.lanitis@cyens.org.cy}
]{\inits{AL}\fnm{Andreas} \snm{Lanitis}}
\author[
   addressref={aff1, aff2},
   email={a.kamilaris@cyens.org.cy}
]{\inits{AK}\fnm{Andreas} \snm{Kamilaris}}

%%%%%%%%%%%%%%%%%%%%%%%%%%%%%%%%%%%%%%%%%%%%%%
%%                                          %%
%% Enter the authors' addresses here        %%
%%                                          %%
%% Repeat \address commands as much as      %%
%% required.                                %%
%%                                          %%
%%%%%%%%%%%%%%%%%%%%%%%%%%%%%%%%%%%%%%%%%%%%%%

\address[id=aff1]{%                           % unique id
  \orgname{Pervasive Systems, University of Twente}, % university, etc
%   \street{Waterloo Road},                     %
  %\postcode{}                                % post or zip code
%   \city{Enschede},                              % city
  \cny{Netherlands}                                    % country
}

\address[id=aff2]{%
  \orgname{CYENS Center of Excellence},
%   \street{D\"{u}sternbrooker Weg 20},
%   \postcode{24105}
  \city{Nicosia},
  \cny{Cyprus}
}

\address[id=aff3]{%
  \orgname{Cyprus University of Tecnhology},
%   \street{D\"{u}sternbrooker Weg 20},
%   \postcode{24105}
  \city{Nicosia},
  \cny{Cyprus}
}

%%%%%%%%%%%%%%%%%%%%%%%%%%%%%%%%%%%%%%%%%%%%%%
%%                                          %%
%% Enter short notes here                   %%
%%                                          %%
%% Short notes will be after addresses      %%
%% on first page.                           %%
%%                                          %%
%%%%%%%%%%%%%%%%%%%%%%%%%%%%%%%%%%%%%%%%%%%%%%

%\note{Sample of title note}     % note to the article
%\note[id=n1]{Equal contributor} % note, connected to author

\end{fmbox}% comment this for two column layout

%%%%%%%%%%%%%%%%%%%%%%%%%%%%%%%%%%%%%%%%%%%%%%
%%                                          %%
%% The Abstract begins here                 %%
%%                                          %%
%% Please refer to the Instructions for     %%
%% authors onhttp://www.biomedcentral.com  %%
%% and include the section headings         %%
%% accordingly for your article type.       %%
%%                                          %%
%%%%%%%%%%%%%%%%%%%%%%%%%%%%%%%%%%%%%%%%%%%%%%

\begin{abstractbox}

\begin{abstract} % abstract
\section*{\small{Background}}
Monitoring species distribution is essential for conservation policies, allowing researchers and policy-makers to assess environmental impacts and appropriate preservation strategies. Traditional data collection, including citizen science, provides valuable yet incomplete insights due to practical limitations. Species Distribution Modelling (SDM) addresses these gaps by using occurrence data and environmental variables to predict species distribution at large geographical areas. In this study, we leverage deep learning and data-imputation techniques to enhance SDM accuracy for frogs, using data from the "EY - 2022 Biodiversity Challenge."

\section*{\small{Results}}
We conducted a series of experiments to evaluate the effectiveness of our proposed method for predicting frog distribution. Data balancing techniques significantly improved model performance, reducing the Mean Absolute Error (MAE) from 189 to 29 in the frog counting task. Additionally, feature selection experiments identified the top environmental variables influencing frog occurrence, streamlining model inputs while maintaining predictive accuracy. Our model demonstrated strong generalization capabilities, achieving comparable MAE scores when applied to unseen geographic regions. Using multimodal input sources (e.g., land cover, NDVI), the final ensemble model outperformed individual models, enhancing both frog counting and occurrence classification accuracy. These results show the potential of our approach for precise and scalable predictions across diverse ecological landscapes.

\section*{\small{Conclusions}}
This study developed a multimodal SDM for frog (Anura) that combines diverse data sources, improving upon traditional SDM approaches reliant on single-modality inputs. By integrating image and tabular data in a fusion model, the model achieved better results in both frog counting and habitat classification. %The model’s ability to accurately count frogs and classify presence/absence areas demonstrates the advantages of multimodal learning for ecological predictions. 
Our approach to generating pseudo-absence data enhanced classification accuracy, achieving 84.9\% with an AUC of 0.90, outperforming other methods and affirming the model’s utility for ecological applications. Additionally, the methodology highlights the importance of data balancing and imputation techniques when datasets available are incomplete or partly accurate.

\end{abstract}

%%%%%%%%%%%%%%%%%%%%%%%%%%%%%%%%%%%%%%%%%%%%%%
%%                                          %%
%% The keywords begin here                  %%
%%                                          %%
%% Put each keyword in separate \kwd{}.     %%
%%                                          %%
%%%%%%%%%%%%%%%%%%%%%%%%%%%%%%%%%%%%%%%%%%%%%%

\begin{keyword}
\kwd{Species Distribution Modeling}
\kwd{Frogs}
\kwd{Data Imputation}
\kwd{Pseduo-absence}
\end{keyword}

% MSC classifications codes, if any
%\begin{keyword}[class=AMS]
%\kwd[Primary ]{}
%\kwd{}
%\kwd[; secondary ]{}
%\end{keyword}

\end{abstractbox}
%
%\end{fmbox}% uncomment this for twcolumn layout

\end{frontmatter}

%%%%%%%%%%%%%%%%%%%%%%%%%%%%%%%%%%%%%%%%%%%%%%
%%                                          %%
%% The Main Body begins here                %%
%%                                          %%
%% Please refer to the instructions for     %%
%% authors on:                              %%
%%http://www.biomedcentral.com/info/authors%%
%% and include the section headings         %%
%% accordingly for your article type.       %%
%%                                          %%
%% See the Results and Discussion section   %%
%% for details on how to create sub-sections%%
%%                                          %%
%% use \cite{...} to cite references        %%
%%  \cite{koon} and                         %%
%%  \cite{oreg,khar,zvai,xjon,schn,pond}    %%
%%  \nocite{smith,marg,hunn,advi,koha,mouse}%%
%%                                          %%
%%%%%%%%%%%%%%%%%%%%%%%%%%%%%%%%%%%%%%%%%%%%%%

%%%%%%%%%%%%%%%%%%%%%%%%% start of article main body
% <put your article body there>

%%%%%%%%%%%%%%%%
%% Background %%
%%

\section{Background}
Ecological research is fundamental to preserve the physical environment and understanding the threats and stressors caused by climate change. For this purpose, researchers require tools to monitor the distribution of flora and fauna, providing crucial information on how species are affected by environmental changes, pollution, and other human activities. Such tools are particularly important for the preservation and protection of endangered species as well as to assess and estimate the degree of impact of invasive species in local ecosystems.

Citizen science projects have emerged as a collaborative approach in which researchers and volunteers gather valuable data on species and their occurrences~\cite{Rowley2019, iNaturalist2024, eBird}. Although these projects offer significant insights into the geolocations of species occurrence, they are inherently incomplete due to the limitations faced by citizen scientists~\cite{Wiersma-2024,10.1093/biosci/biw022, SICACHAPARADA2021100446, WEISER2020108411}, such as large geographic regions, time constraints, the challenge of covering every location due to insufficient resource and limited expertise of citizen scientists, increasing the likelihood of misidentification and false reporting. To address these gaps, simpler approaches like interpolation are traditionally used, where estimates for unsurveyed regions are made based on nearby data points. However, such methods often lack precision and fail to account for the complex environmental variables influencing species distribution. To overcome these limitations, the concept of Species Distribution Model (SDM) has been introduced. 

Species Distribution Models can be defined as "a quantitative, empirical model of species-environment relationships developed using geo-location of species data and the environmental features that affect those species distributions. The methodology or techniques used to develop such Species Distribution Models are called Species Distribution Modelling"~\cite{ELITH2017}. SDMs are an important tool that contributes greatly to biodiversity research, which in turn supports ecological conservation~\cite{sdm-app}. These models provide a measurable framework to explain the relationship between input variables or covariates (which may include environmental features, climatic factors, or remotely sensed images of landscapes and habitats) and the distribution of a species. The primary function of SDMs is to map the spatial distribution of a species based on occurrence data obtained from multiple sources. One of the critical applications of SDMs is in the study of bio-indicators—organisms that provide valuable insights into the health of an ecosystem. These organisms, which can include animals, plants, and microorganisms, are particularly sensitive to changes in environmental conditions. Their presence, absence, or well-being can reveal much about the quality of the environment. Frogs (Anura) are especially significant as bio-indicators due to their heightened sensitivity to environmental shifts,such as changes in water quality, temperature, and pollution levels~\cite{iman2020MarshF, kurnianto2024assessing}. The study of frogs using SDMs is crucial because it allows researchers to track and predict their distribution in relation to environmental changes, thereby offering a deeper understanding of broader ecological impacts.

In this work, we leverage a multimodal deep learning model that utilizes data from multiple sources to develop a robust species distribution model (SDM). The motivation for this approach arises from the promising results that multimodal learning has demonstrated in various applications, including SDMs~\cite{8269806, 9068414, 10.5555/3104482.3104569, article_sen}. Specifically, this study explores the potential of multimodal learning to create more accurate and comprehensive SDMs, using the dataset provided by the '2022 Biodiversity Challenge'~\cite{EY2022}, with a focus on species such as frogs.~\cite{EY2022}.

\subsection{Related Work}
\label{RelWork}
In SDM, the modeling approach to be adopted is closely tied to the type of data available. In contrast, presence-absence data includes both occurrences and absences, and each type of data presents unique challenges and opportunities for SDM modeling purposes. Presence-only data can be limited due to its lack of absence information, while presence-absence data provides a more comprehensive foundation for analysis. The characteristics of the datasets provided for SDM studies significantly influence the effectiveness of various modeling techniques. In this study, we review the existing literature and categorize it into three main approaches: Statistical Methods, Machine Learning Methods, and Multi-modal Deep Learning Methods.

\subsubsection{Statistical Methods}

Statistical methods have long served as a foundational approach in SDM, particularly for presence-only data. Various models, such as MaxEnt~\cite{10.1145/1015330.1015412, PHILLIPS2006231}, estimate species distributions by calculating maximum entropy. While MaxEnt is widely used due to its robustness, it is limited by its reliance on presence-only data, which can introduce significant biases and lead to overestimations in species' ecological niches. Generalized linear models, like logistic regression, have also been employed effectively in SDMs. Manel et al.~\cite{MANEL1999337} demonstrated the effectiveness of logistic regression in predicting species distributions, though they noted the tendency for overfitting, especially when predictor variables are correlated by chance rather than reflecting true ecological relationships.

\subsubsection{Machine Learning Methods}

As machine learning techniques have evolved, they have increasingly been applied to SDMs, offering the ability to capture complex, non-linear relationships between environmental factors and species presence. For instance, random forests and support vector machines have shown promise in improving predictive performance. Cutler et al.~\cite{14_https://doi.org/10.1890/07-0539.1} utilized random forests to model species distributions, outperforming traditional methods like logistic regression and linear discriminant analysis. However, the complexity of machine learning models can make the interpretation of ecological relationships challenging.

Deep learning methods have further advanced the field by effectively handling large and complex datasets. Although early neural networks demonstrated improved predictions, deep neural networks (DNNs) with multiple hidden layers have shown even greater efficacy in capturing intricate relationships among variables~\cite{10_botella:hal-01834227}. Despite their potential, DNNs are prone to overfitting, particularly with smaller datasets, and require careful hyperparameter tuning~\cite{62_Shiferaw2019}. Convolutional neural networks (CNNs) represent a specialized form of deep learning that excels in processing spatial data. Deneu et al.~\cite{15_10.1371/journal.pcbi.1008856} proposed a CNN-based SDM, leveraging spatial environmental tensors to model complex ecological niches. Their findings indicate that CNNs can significantly improve predictions, particularly when occurrence data is limited.

\subsubsection{Multimodal Deep Learning Methods}
While deep learning-based SDMs have shown promising performance, they typically rely on a single modality, either environmental variables or high-resolution satellite images. This limitation can result in missing critical spatial patterns when only environmental data—such as temperature, soil type, and humidity—is considered. Conversely, using only satellite images neglects valuable climatic information, adversely affecting model performance. To address these challenges, multimodal learning can effectively handle the heterogeneous nature of input data in SDMs. A key aspect of multimodal learning is the fusion of representations from different modalities, which significantly influences the overall performance of the model. 

Multimodal learning has gained traction in various applications, allowing deep neural networks to learn features from multiple data modalities simultaneously. Notable applications include video and image captioning~\cite{68_Vinyals2014ShowAT, 71_Xu_2016_CVPR}, image generation from text~\cite{xu2017attnganfinegrainedtextimage, 74_yan2016attribute2imageconditionalimagegeneration}, and speech recognition, which combine audio and visual data~\cite{20_865479}. However, the application of multimodal learning in the field of SDMs remains limited.

Deneu et al.~\cite{16_deneu:hal-02989084} demonstrated a multimodal approach by fusing input data at the initial stage before feeding it into a convolutional neural network (CNN). However, this method may limit the model's ability to capture unique information from diverse modalities, particularly when those modalities differ significantly.Seneviratne~\cite{60_article} incorporated multimodal imagery in habitat prediction for 30,000 species by training a ResNet50 model on RGB images and subsequently adding altitude images. This architecture utilized a separate branch for altitude features, merging higher-level features at the end. This design allows for more nuanced learning, as it adapts better to different image modalities, leading to lower error rates compared to unimodal structures.

Similarly, Zhang et al.~\cite{su142114034} explored two approaches for multimodal learning in SDMs. Their first approach featured a two-branch structure: one branch employed a ResNet architecture for remote sensing RGB images, while the other processed 27 environmental variables using a fully connected layer. The concatenated feature vectors were then processed through additional layers to obtain final predictions. Their second approach utilized a Swin Transformer based on an attention mechanism, implementing multiple fusion methods (pre-fuze, post-fuze, mid-fuze, and feature addition) to optimize model performance. The results indicated that the pre-fuze method achieved the highest accuracy among the tested approaches.

% Our work contributes to this field by proposing a range of data balancing techniques, including Adaptive Oversampling and a customized loss function, effectively balancing the dataset and ensuring adequate representation for minority classes. We introduce a pseudo-absence data imputation method that considers critical factors influencing the accuracy of SDMs, enhancing the reliability of absence data essential for accurate predictions. Furthermore, we develop a multimodal deep learning architecture utilizing a late fusion strategy with a pretrained backbone for both classification and regression tasks. This architecture leverages the strengths of different data modalities to improve model performance, addressing the limitations identified in existing research.

\subsection{Our Contribution}
Based on the comprehensive literature review, our contributions can be summarized as follows:

\begin{enumerate}
    \item We propose a range of data balancing techniques, including Adaptive Oversampling and a Customized Loss Function. These methods effectively balance the dataset, ensuring that it captures the diversity across all classes while providing adequate representation for minority classes.
    
    \item We introduce a pseudo-absence data imputation method that takes into account several critical factors influencing the accuracy of SDMs. This approach enhances the reliability of absence data, which is crucial for accurate predictions.
    
    \item We develop a multimodal deep learning architecture utilizing a late fusion strategy with a pretrained backbone for both classification and regression tasks. This architecture leverages the strengths of different data modalities to improve model performance.
\end{enumerate}

The diverse approaches proposed in this work have demonstrated effectiveness, as evidenced by our results, which achieve accuracy levels comparable to the challenge winner.

\section{Methodology}
\subsection{Dataset}
In this experiment, we used the EY Frog Counting Challenge dataset~\cite{EY2022}. The challenge invites researchers to develop computational models for predicting frog occurrence and counts at specific locations using multiple datasets. The challenge provided a list of predictor variables, known as environmental covariates, from which we selected the most relevant ones for our analysis, along with the ground truth, which is the frog occurrence data. We provide a detailed description of the datasets and the predictor variables in the following sub-sections.

\subsubsection{Environmental Covariates (Predictor Variables/Covariates)}
\label{covarates}
The environmental covariates used in this study were downloaded from the Microsoft Planetary Computer Portal~\cite{Microsoft2023}, an open-source platform that grants access to a diverse array of high-resolution geospatial data. We selected the covariates that affect the ecological conditions influencing frog populations. The datasets employed as predictor variables include:

\begin{itemize}
    \item Sentinel-2 Level-2A~\cite{Sentinel2Level2A}: This dataset provides high-resolution satellite imagery across 13 spectral bands, with resolutions ranging from 10m to 60m. For this analysis, we utilized the Red, Green, Blue (RGB), and Near-Infrared (NIR) bands, which are instrumental in assessing vegetation health and land cover characteristics.
    \item JRC Global Surface Water (GSW)~\cite{Pekel2016}: This dataset offers comprehensive information regarding the occurrence, seasonality, and transitions of surface water on a global scale. It is vital for understanding the aquatic habitats that influence frog populations.
    \item Esri 10-meter Land Cover~\cite{Karra2021}: This dataset categorizes land cover types at a spatial resolution of 10 meters, including water bodies, forests, grasslands, and agricultural areas. This classification is crucial for determining the habitat preferences of frogs and their distribution patterns.
    \item Copernicus DEM GLO-90~\cite{CopernicusDEM}: The Copernicus Digital Elevation Model provides detailed elevation data, which is particularly relevant for assessing how species distributions may shift in response to climate change, such as upward movements of amphibians due to rising temperatures.
    \item  TerraClimate~\cite{Abatzoglou2018}: This dataset provides high-resolution monthly climate and climatic water balance data for global terrestrial surfaces, derived through climatically aided interpolation. For this study, we selected temperature (both minimum and maximum), precipitation, Palmer Drought Severity Index, vapor pressure, and soil moisture. These factors are known to significantly affect frog habitat and distribution, as frogs rely on specific temperature ranges and moisture levels for survival and reproduction.
\end{itemize}

Apart from the raw data obtained from the datasets mentioned above, we included two derived indices using Sentinel-2 Level-2A data: the Normalized Difference Vegetation Index (NDVI) and the Normalized Difference Water Index (NDWI), which measure vegetation and water levels, respectively.
\begin{itemize}
    \item Normalized Difference Vegetation Index (NDVI) \cite{Rouse1973MonitoringVS}: This index provides a quantifiable measure of vegetation in a particular location and is a significant factor in assessing habitat suitability for frogs. NDVI is calculated using the Equation \ref{eq:NDVI}.
    % \begin{equation}
    %     \text{NDVI} = \frac{\text{NIR} - \text{Red}}{\text{NIR} + \text{Red}} \quad 
    % \end{equation}

    \item Normalized Difference Water Index (NDWI) \cite{doi:10.1080/01431169608948714}: NDWI provides an index that indicates the surface water content, which also significantly influences frog habitat. NDWI is calculated using the Equation \ref{eq:NDVI}.
    % \begin{equation}
    %     \text{NDWI} = \frac{\text{Green} - \text{NIR}}{\text{Green} + \text{NIR}} \quad
    % \end{equation}
\end{itemize}

\begin{align}
    \text{NDVI} &= \frac{\text{NIR} - \text{Red}}{\text{NIR} + \text{Red}} \quad,   \hspace{1cm}
    \text{NDWI} = \frac{\text{Green} - \text{NIR}}{\text{Green} + \text{NIR}} \quad \label{eq:NDVI}
\end{align}

 where NIR, Green and Red are the spectral reflectances of the Near Infrared, Green and Red channels, respectively.

\label{DataColl}
\subsubsection{Frog Occurrence Dataset (Target Variable)}
The frog occurrence dataset provided by the challenge~\cite{EY2022}, which serves as the target variable in this analysis, was compiled from citizen science initiatives across three countries: Australia, South Africa, and Costa Rica. The Australian data originates from the FrogID project~\cite{Rowley2019}, while the South African and Costa Rican datasets are part of iNaturalist Research-grade observations~\cite{iNaturalist2024}. These projects aim to enhance biodiversity knowledge by aggregating observations made by individuals in various locations. The initial dataset included comprehensive information, such as species names, geographical coordinates of observations, and timestamps. This raw data provides the foundation for analyzing frog populations and their habitat requirements, enabling the development of an effective SDM.

\subsection{Data Preprocessing}
To make the data usable for such predictions, several transformations are applied, including grid creation, and data balancing. Each of these preprocessing steps is essential for ensuring that the dataset is structured appropriately for analysis and that it meets the requirements of the predictive model.

In this subsection, we will discuss each of these transformations in detail:
\subsubsection{Grid Creation}
\begin{figure}[ht]
\centering
\includegraphics[width=0.96\textwidth]{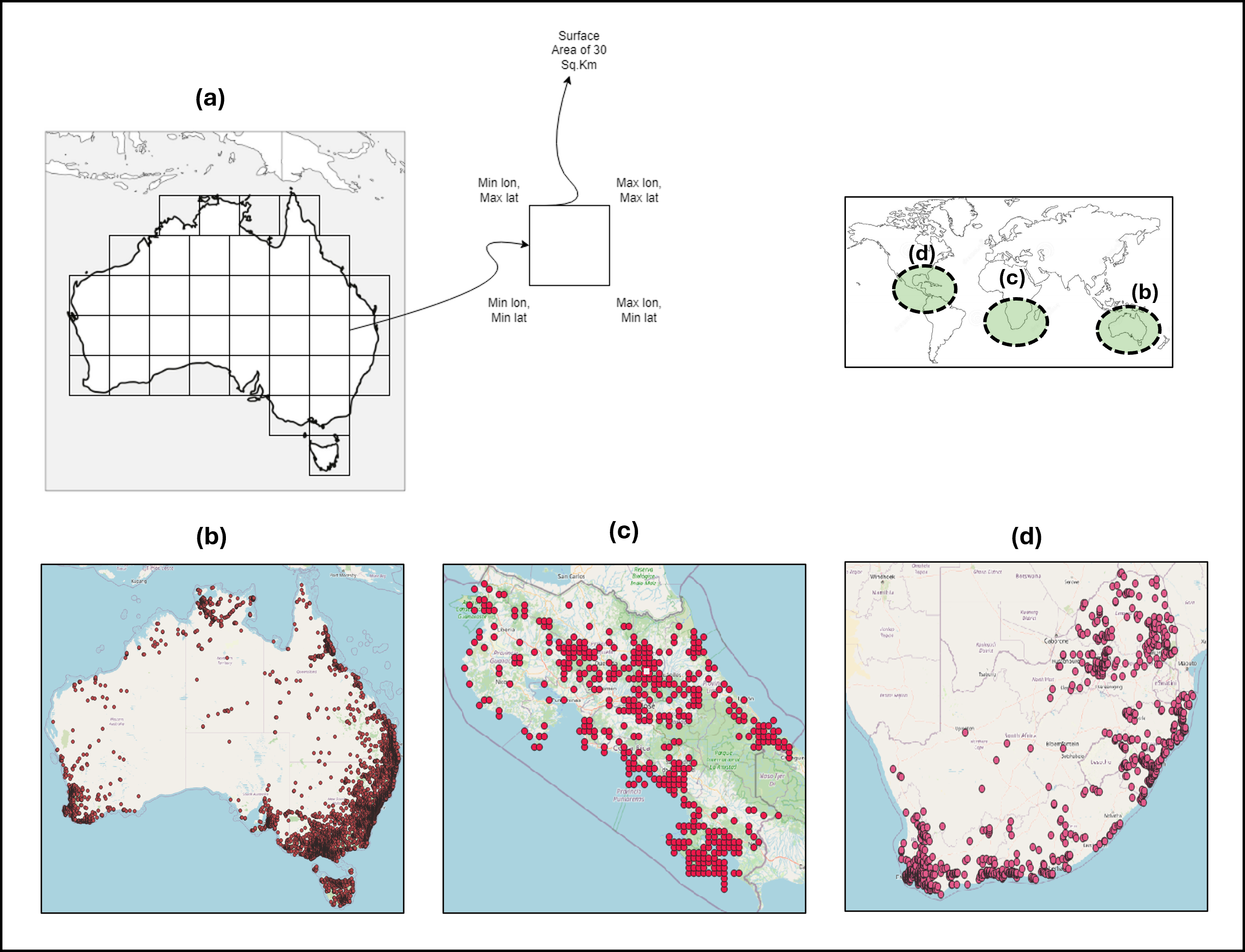}
\caption{(a) Grid creation for calculating frog density across three countries using 30 km² cells. (b,c,d) Frog presence points in Costa Rica, Australia, and South Africa.}
\label{fig:grid-creation}
\end{figure}

To calculate frog density across all three countries, we create grids of 30 square kilometers, partitioning each country into several grid cells. These grids are defined by bounding box coordinates (minimum latitude, minimum longitude, maximum latitude, maximum longitude). Once the grids are established, we obtain the frog count for each grid by iterating through each cell and subsetting the frog presence points from the original dataset. This grid structure is also used to prepare the covariates defined in subsection~\ref{covarates}. An illustration of the grid creation for Australia is shown in Fig~\ref{fig:grid-creation}, while the presence points for Costa Rica, South Africa, and Australia are visualized also in Fig~\ref{fig:grid-creation} using QGIS software.

\subsubsection{Data balancing using Adaptive Oversampling}
\label{data-balance1}
\begin{figure}[ht]
\centering
\includegraphics[width=.80\textwidth]{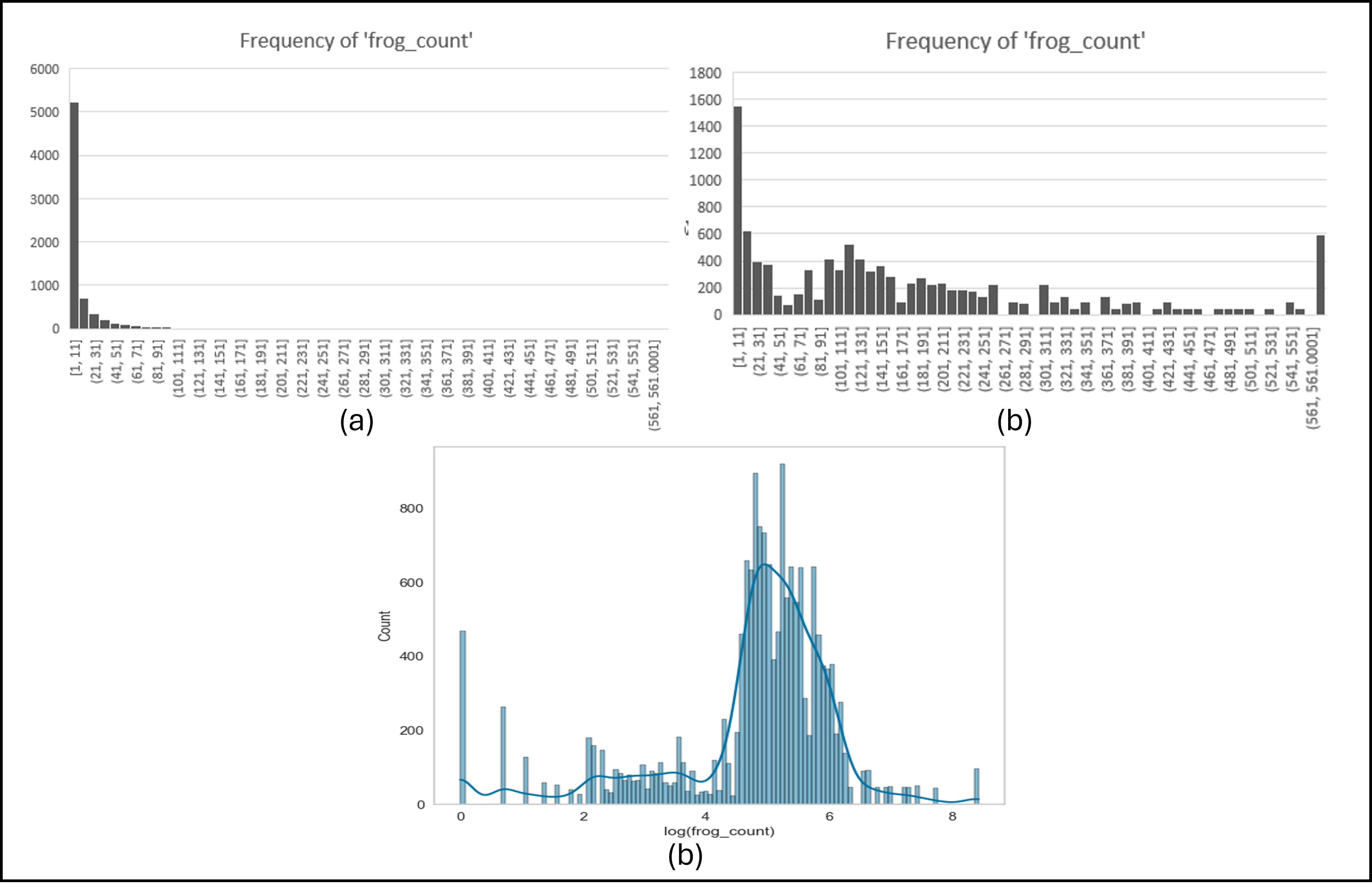}
\caption{Histogram of Unbalanced, Balanced and Log transformed frequency of frog occurrence.}
\label{fig:balance_vs_unblance}
\end{figure}

In our study, the frog count serves as the target variable, and as illustrated in Fig~\ref{fig:balance_vs_unblance}(a). In the dataset there is a significant imbalance, with frog counts predominantly concentrated between 1 and 10. This skewed distribution poses challenges for effective modeling, as the minority counts may not be adequately represented, potentially leading to biased predictions. To address this issue, we employ a technique known as Adaptive Oversampling. This approach enhances the representation of the minority class by smartly augmenting the dataset rather than simply duplicating instances, which can diminish the dataset's diversity and variability. By leveraging clustering techniques, we aim to restore the original variability of the dataset while achieving a more balanced representation of frog counts.

In this method, the algorithm initially identifies the minority class based on the frequency of occurrences. Instead of merely duplicating instances from this class, Adaptive Oversampling employs K-means clustering to generate new, unique samples. This process begins with feature selection from predictor variable (here, TerraClimate) and by defining the number of clusters. TerraClimate was chosen as the primary dataset because it provides high-resolution, ecologically relevant climatic variables, such as temperature, precipitation, and soil moisture, which are critical factors influencing species distribution and habitat suitability. These variables directly affect the physiology, reproduction, and survival of frogs, making them particularly relevant for our study. While satellite imagery captures fine-scale spatial features, its high dimensionality can complicate clustering and may not directly reflect the climatic and ecological drivers of species distributions. By focusing on TerraClimate for the clustering step, we could reduce computational complexity while still generating meaningful samples.

The K-means algorithm clusters similar data points together, allowing for the selection of unique representatives from each cluster. By combining these unique instances with the original dataset, Adaptive Oversampling effectively balances the representation of minority classes without sacrificing the variability of the dataset. This method not only enhances the model's ability to learn from the data but also contributes to more robust and generalizable predictive performance. After applying Adaptive Oversampling, we obtained a better-balanced dataset, as shown in Fig~\ref{fig:balance_vs_unblance}(b).

\begin{algorithm}
    \caption{Balance Dataset with Oversampling and K-means Clustering}
    \begin{algorithmic}[1]
        \State \textbf{Input:} Dataset $data$
        \State \textbf{Output:} Balanced dataset $balanced\_data$
        
        \State $data \gets \text{load\_dataset()}$
        \State $frequency \gets \text{calculate\_frequency}(data)$
        \State $minority\_class \gets \text{identify\_minority\_class}(frequency)$

        \Function{oversample}{$data$, $minority\_class$, $n\_samples$}
            \State $oversampled\_data \gets []$
            \For{each $instance$ in $minority\_class$}
                \For{$i \gets 1$ to $n\_samples$}
                    \State $oversampled\_data.\text{append}(instance)$
                \EndFor
            \EndFor
            
            \Return $oversampled\_data$
        \EndFunction

        \State $features \gets \text{select\_features}(data, terraclimate\_data)$

        \State $n\_clusters \gets \text{define\_number\_of\_clusters}()$

        \State $centroids, cluster\_labels \gets \text{kmeans\_clustering}(features, n\_clusters)$

        \State $unique\_instances \gets []$
        \For{cluster in range($n\_clusters$)}
            \State $cluster\_points \gets \text{get\_points\_in\_cluster}(data, cluster\_labels, cluster)$
            \State $unique\_instance \gets \text{select\_unique\_instance}(cluster\_points)$
            \State $unique\_instances.\text{append}(unique\_instance)$
        \EndFor

        \State $balanced\_data \gets \text{combine\_data}(data, unique\_instances)$

        \State \text{output\_balanced\_dataset}(balanced\_data)
    \end{algorithmic}
\end{algorithm}

\subsubsection{Data Balancing using Custom Loss Function}
\label{data-balance2}
In addition to the frequency-based imbalance in frog counts, the dataset exhibits an imbalance across countries. Among the three countries represented, Australia accounts for approximately 82\% of the data, while South Africa contributes 11\% and Costa Rica only 7\%. This disparity can result in biased models that perform poorly on the minority classes. To address this issue, during the training phase, a weighted loss function was introduced. This approach involves assigning a higher weight to the minority classes (Costa Rica and South Africa) and a lower weight to the majority class (Australia). The weights for each class are intuitively computed using Equation~\ref{class_weight}:

\begin{equation}
    \text{Weight of class } x = \frac{\text{Total number of samples}}{\text{Number of samples in class } x} \times \text{Number of Classes} \quad 
    \label{class_weight}
\end{equation}

These class weights are then multiplied by the original loss value with the addition of, as shown in Equation~\ref{loss_fun}:

\begin{equation}
    \text{Total loss} = (\text{Weight of class } (x) \times \text{loss value}) + \text{regularization loss}
\label{loss_fun}
\end{equation}

This custom loss function is applied during the training phase to help mitigate the impact of class imbalance, thereby enhancing the model's ability to learn effectively from minority class examples. By integrating this technique into the training process, we aim to achieve a more equitable performance across all classes, ultimately leading to a more robust predictive model.

\subsubsection{Data Balancing using Log Transformation}
\label{data-balance3}
Skewness, which measures the asymmetry of data distribution, is a common issue that can adversely affect model performance. When the data is not normally distributed, model performance tends to suffer. In this study, the data is positively skewed, as evidenced by the long tail in the positive direction, as shown in Fig~\ref{fig:balance_vs_unblance}(b). To address this, a log transformation is applied to the dependent variable, effectively reducing skewness. Refer to Fig~\ref{fig:balance_vs_unblance}(c) for the histogram of frog counts after transformation. It is essential to note that predictions from the model will be in log scale, so the inverse transformation must be applied for accurate interpretation.

\subsubsection{Data Balancing using Image Augmentation} 
To enhance dataset variability and uniqueness, various image augmentation techniques are utilized. These techniques not only help balance the dataset but also make the model more generalized and prevent overfitting. The following table summarizes the different image augmentation methods employed:

\begin{table}[ht]
    % \centering
    \caption{Image Augmentation Techniques}
    \begin{tabular}{l l l}
        \hline
        \textbf{Technique}  & \textbf{Description} \\ 
        \hline
        Horizontal Flip  & Probability: 0.5 \\ 
        % \hline
        Rotation & Angle: Randomly chosen between -10 and 10 degrees \\ 
        % \hline
        Scaling & Factor: Randomly sampled between 0.6 and 1.4 \\ 
        % \hline
        Resizing & Randomly resizes \\ 
        \hline
    \end{tabular}
    \label{tab:image_augmentation}
\end{table}

\subsection{Imputation of Pseudo-absence Data} A significant challenge in species distribution modeling lies in obtaining absence data. Collecting true absence data for any species poses significant challenges. Unlike presence data, which can be more readily gathered through direct observation or surveys, absence data is often complicated by various factors. For instance, a species may not be observed at a particular site, but this does not necessarily indicate its absence. Factors such as human error during observations, local extirpation due to anthropogenic influences, or migratory patterns can all contribute to a species not being detected, even in environments that are otherwise suitable for its presence. These complexities necessitate innovative approaches to generate reliable absence data, which is crucial for understanding species distribution and informing conservation efforts.

To address the challenge of obtaining reliable absence data, the proposed pseudo-data imputation technique outlines a procedure for generating pseudo-absence data, carefully considering several critical factors that may influence the accuracy of SDM. One key consideration is the ratio of pseudo-absence points to presence points; an imbalanced dataset can significantly skew results and affect model performance. Additionally, the selection of covariates used to filter the data—such as geographical extent and land cover types—plays a vital role in ensuring that the generated pseudo-absence data accurately reflects potential habitats for the species in question.

% \subsubsection*{Imputation Method}
The proposed imputation method divides the study area into 30 km² grid. The grid cells containing known presence data are identified, while the remaining grid cells are classified as potential pseudo-absence points. Following this, presence points are selected based on a specified distance from these potential pseudo-absence points, utilizing the Haversine-formula~\cite{haversine_wikipedia} to calculate distances based on geographic coordinates. The distance thresholds are tailored for three distinct countries: 10 km for Australia, 20 km for South Africa, and 28 km for Costa Rica, reflecting the varying densities of presence data across these regions. These distance thresholds were determined based on the density of presence data points in each country. Smaller thresholds were used for regions with higher data densities, while larger thresholds were applied to sparser regions to ensure a balanced selection of pseudo-absence points.

Subsequently, land cover types at both presence points and the newly identified potential pseudo-absence points are analyzed using the Esri 10-meter land cover dataset. Points that share the same land cover type as the presence points and fall within the threshold distance are classified as potential pseudo-absence points. The land cover type was prioritized for this task because: 1) it represents similar landscapes as the presence points, making it more likely that the location was visited by citizen scientists, yet frogs may not have been observed; and 2) similar land cover types are expected to exhibit comparable environmental and climatic conditions. Climate data was not included, as land cover alone was sufficient to capture these similarities. This methodology not only enhances the reliability of the generated absence data but also provides a more nuanced understanding of the species' distribution patterns, ultimately aiding in more effective conservation strategies.

\begin{algorithm}
\caption{Pseudo-Absence Data Generation}
\begin{algorithmic}[1]
\State \textbf{Input:} Study Area, Presence Points, Distance Thresholds (for each country)
\State \textbf{Output:} Pseudo-Absence Points

\Procedure{GeneratePseudoAbsenceData}{}
    \State Divide the Study Area into 30 km² grid cells
    \State Identify grid cell with Presence Points
    \State Remaining grid cell $\gets$ Potential Pseudo-Absence Points

    \For{each Presence Point}
        \State Calculate distance to all Potential Pseudo-Absence Points using Haversine formula
        \For{each country}
            \If{country == Australia}
                \State Set Distance Threshold $X \gets 10 \text{ km}$
            \ElsIf{country == South Africa}
                \State Set Distance Threshold $X \gets 20 \text{ km}$
            \ElsIf{country == Costa Rica}
                \State Set Distance Threshold $X \gets 28 \text{ km}$
            \EndIf

            \State Select points at distance $X$ from Presence Points
        \EndFor
    \EndFor

    \State Retrieve Land Cover Types for Presence Points and Pseudo-Absence Candidates
    \For{each Candidate Point}
        \If{Land Cover Type of Candidate == Land Cover Type of Presence Point}
            \State Classify Candidate Point as Pseudo-Absence Point
        \EndIf
    \EndFor

    \State \textbf{Return} Pseudo-Absence Points
\EndProcedure

\end{algorithmic}
\end{algorithm}

\subsection{Model}
This section describes the model architecture used to build the SDM, subdivided into two parts according to the nature of the problems addressed. The first part details the architecture utilized for the frog presence/absence classification task, while the second part focuses on the regression task for counting frogs.
\begin{figure}
\centering
\includegraphics[width=.95\textwidth]{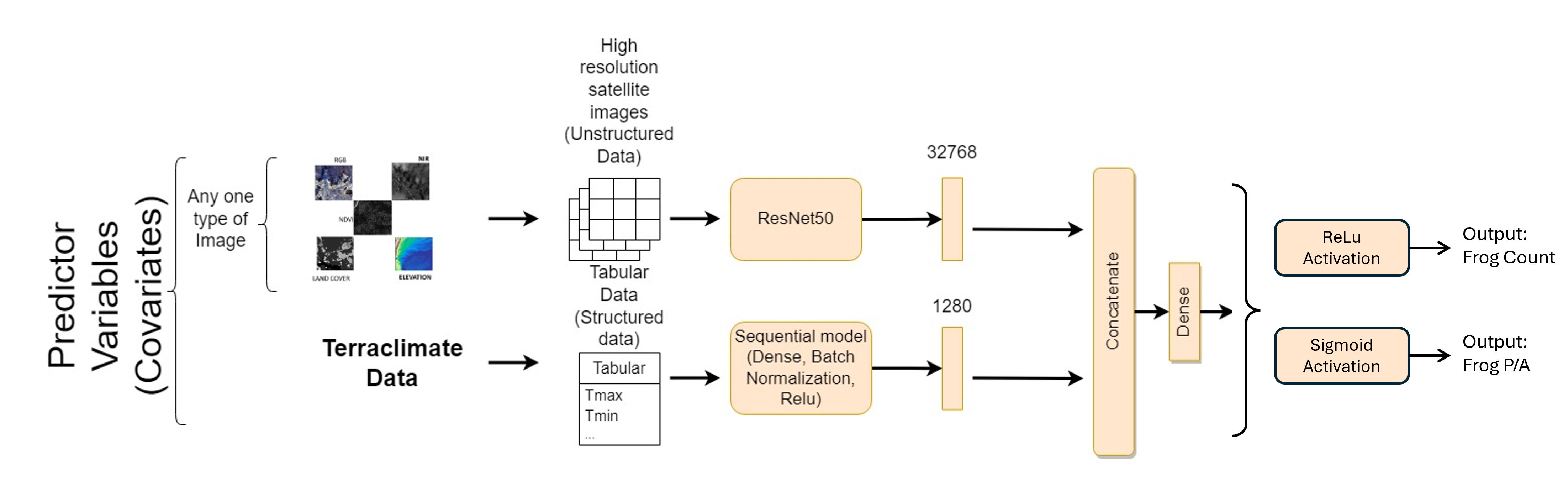}
\caption{Architectures of the multimodal learning models for frog presence/absence classification and frog counting regression, presented together for clarity. Both models combine highresolution satellite images and TerraClimate data as inputs. The classification model uses a sigmoid activation function to predict frog presence/absence, while the regression model employs a ReLU activation function to output frog count predictions.}
\label{fig:Block_diagram}
\end{figure}

\subsubsection{Frog Presence/Absence Classification}
The classification task employs a multimodal learning approach, integrating input data from multiple modalities to predict the presence or absence of frogs. The fusion module facilitates this combination of different models, resulting in a unified representation of the different modalities used in the classification process. In this work, late fusion is employed to enable each modality to maintain its own neural network for feature extraction, effectively capturing the unique characteristics of the input data before combining them in latent space.

The specific architecture (Model A) used is illustrated in Fig~\ref{fig:Block_diagram}. This fusion model integrates information from two distinct input sources: high-resolution satellite images and numerical/tabular data, specifically Terraclimate data. These inputs constitute the predictor variables or covariates. The model contains two branches. Input images are fed into a ResNet50~\cite{He2015DeepRL} model pre-trained on the ImageNet dataset~\cite{5206848}. ResNet50 was chosen as the backbone model for its proven ability to extract robust and meaningful features from high-resolution images in a variety of computer vision tasks. The output from the ResNet50 model consists of features extracted from the image inputs, which are then flattened to yield 32,768 features. Meanwhile, the sequential model produces a flattened array of 1,280 features from the numerical data. The extracted features from both modalities are concatenated to form a combined feature vector of length 34,048. This vector is subsequently passed through a dense layer, followed by a softmax activation function to predict the presence or absence of frogs.

We trained three models (shown in Fig~\ref{fig:Block_diagram}) with different input combinations, they are:
\begin{itemize}
    \item Model A: High-resolution satellite imagery and TerraClimate data. 
    \item Model B: Land cover patches and TerraClimate data.
    \item Model C: NDVI data TerraClimate data. 
\end{itemize}

\subsubsection{Frog Counting Task Regression}
The architecture used for the counting task closely resembles that of the classification task, with the key difference being the use of a ReLU activation function at the output layer. This model predicts the count of frogs present at a given location. Similar to the classification problem, this task was implemented for three sets of data: RGB \& Terraclimate, land cover \& Terraclimate, and NDVI \& Terraclimate.

\subsubsection{Loss Function}

For the frog presence/absence classification task, Binary Cross-Entropy Loss is employed. As this is a binary classification problem, the Binary Cross-Entropy Loss function quantifies the difference between the predicted probabilities and the actual binary labels. The mathematical formulation is given in Equation~\ref{bce}:

\begin{equation}
\text{Binary Cross-Entropy Loss} = -\frac{1}{N} \sum_{i=1}^{N} \left( y_i \cdot \log(p_i) + (1 - y_i) \cdot \log(1 - p_i) \right)  \quad 
\label{bce}
\end{equation}

where \( N \) represents the number of data points, \( y_i \) denotes the true value, and \( p_i \) indicates the predicted probability.

For the frog counting task, the Mean Squared Logarithmic Error (MSLE) is utilized as the loss function. Given that the target variable encompasses a wide range of continuous values, MSLE treats small differences between actual and predicted values similarly to larger differences. The MSLE is calculated using the formula provided in Equation~\ref{msle}:

\begin{equation}
\text{MSLE} = \frac{1}{N} \sum_{i=1}^{N} \left( \log(y_i + 1) - \log(\hat{y}_i + 1) \right)^2  \quad 
\label{msle}
\end{equation}

where \( N \) is the number of data points, \( y_i \) represents the true value, and \( \hat{y}_i \) denotes the predicted value.

In addition to the loss functions, regularization techniques are employed to prevent overfitting during training. Specifically, L2 regularization\ref{loss_fun}, also known as ridge regression, is applied, where the square of the magnitude of the coefficients is added as a penalty to the loss function.

\subsection{Optimizer and Learning Rate}
The optimizer utilized for training the model is Adam. Although Adam autonomously manages learning rate optimization for each parameter, a learning rate scheduler is implemented to incorporate a warm-up phase. This warm-up phase facilitates a smooth transition from the initial learning rate to the target value, enhancing the training process.

\subsection{Training, Testing and Evaluation Data}
The dataset was divided into training and testing subsets in an 80:20 ratio. The training subset was used to optimize the model parameters, while the testing subset was reserved for evaluating the model's performance on unseen data. Additionally, the model's generalization ability was assessed using validation data provided by the challenge~\cite{ey_wavespace_madrid}. This comprehensive approach ensured a thorough evaluation of the model's performance in real-world scenarios.
The validation data provided by the challenge consisted of larger grid cells sized at 225 km², compared to the 30 km² grid cells used during training. To address this difference, two methods were tested: resizing the validation grid cells to match the 30 km² resolution and applying a sliding window approach to divide the larger grid cells into smaller patches. Resizing was found to be more effective, as it preserved spatial consistency and better aligned with the model's training scale. Based on this finding, the resizing method was adopted for all subsequent experiments.

\subsubsection{Evaluation Metrics}
\subsubsection*{Frog Occurrence - Classification Task}
For the binary classification task, accuracy is chosen as the evaluation metric. Accuracy is calculated as the ratio of correctly predicted instances to the total instances in the dataset. However, accuracy is heavily dependent on the chosen threshold, so to minimize the influence of this threshold, the area under the ROC curve (ROC AUC) is also used as a secondary evaluation metric.

\subsubsection*{Frog Counting - Regression Task}
For the regression task of predicting frog counts, Mean Absolute Error (MAE) is chosen as the evaluation metric. MAE quantifies the average absolute difference between the actual and predicted values. In simple terms, MAE indicates how much the predicted values deviate from the actual counts; a larger MAE signifies poorer model performance. MAE was also used to evaluate the performance in the EY-Frog Challenge, allowing for a direct comparison between the proposed model and the submitted results. The MAE is calculated using the formula given in Equation~\ref{mae_equ}:
\begin{equation}
\text{MAE} = \frac{1}{n} \sum_{i=1}^{n} |y_i - \hat{y}_p|  \quad 
\label{mae_equ}
\end{equation}
where \( n \) is the number of observations, \( y_i \) represents the true values, and \( \hat{y}_p \) denotes the predicted values.

\section{Results and Discussion}
\label{results}

This section presents the results of multiple experiments conducted to evaluate the proposed method. We analyze the effectiveness of different data balancing techniques, assess feature importance, and evaluate how well the model generalize to geographical regions. Following these experiments, we present the performance of the frog counting task (regression) and the frog occurrence task (classification) for different input data sources. Results from various pseudo-absence data generation techniques are also presented, comparing them to the proposed method. Finally, we compare both the regression and classification results with those of the challenge's winner to further validate the effectiveness of our approach.

\subsection{Comparison of Data Balancing Results}
Balancing the dataset is a crucial yet often underexplored aspect of building SDM. In this study, we propose a range of data balancing techniques and analyze the impact of the balanced dataset by comparing the performances of models trained on both the original imbalanced dataset and the balanced dataset. 

\begin{figure}
\centering
\includegraphics[width=.9\textwidth]{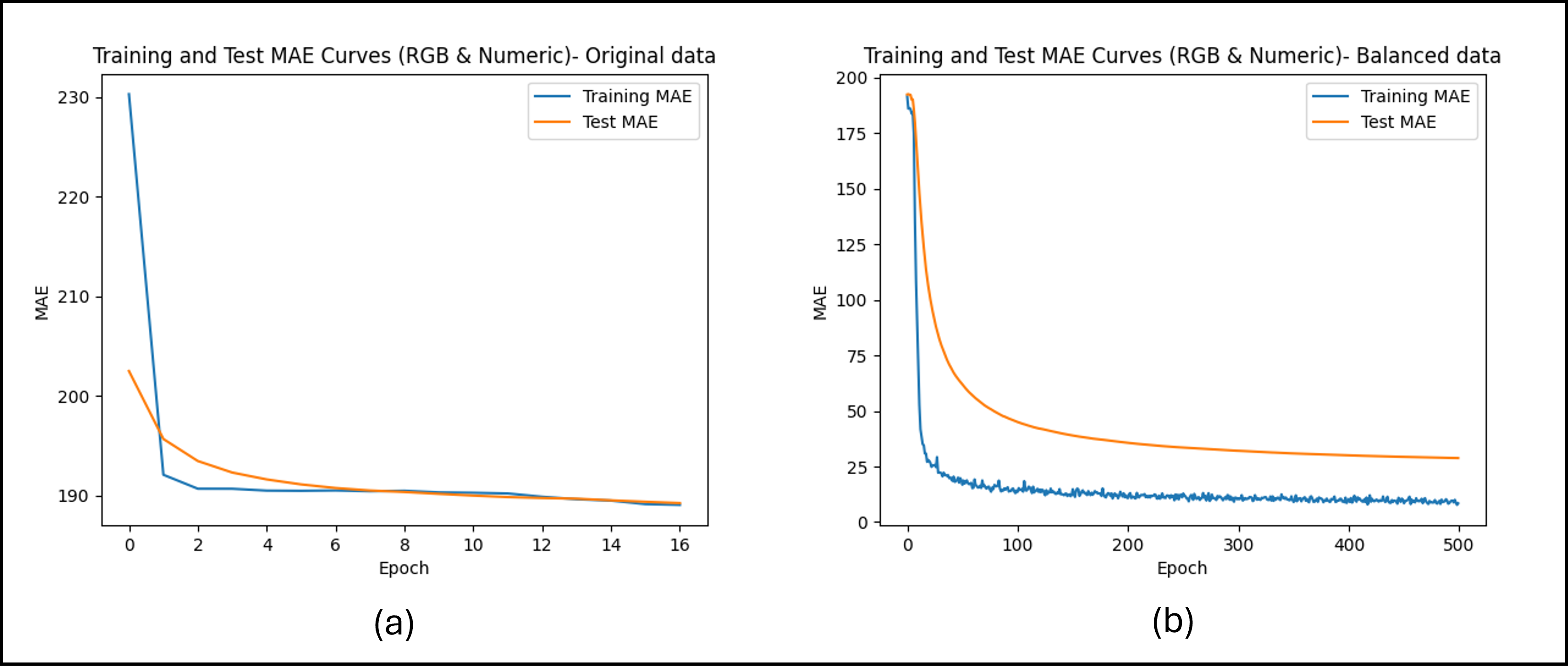}
\caption{(a) Train and Test MAE curves for Imbalanced data, (b) Train and Test MAE curves for balanced data}
\label{fig:train_curve}
\end{figure}

The results obtained before and after balancing the dataset are shown in Table~\ref{tab:mae_comparison}. For comparison purposes, only the Model-A (RGB and Numeric data) is presented. The Mean Absolute Error (MAE) for the model trained on the imbalanced data shows a high value of around 190 for both training and testing. From Fig~\ref{fig:train_curve}(a), it can be observed that after epoch 2, the learning curve began to saturate and did not decrease further, indicating the model's inability to learn effectively. After balancing the dataset, following the techniques described in subsection~\ref{data-balance1},~\ref{data-balance2},and~\ref{data-balance3}, the model achieved a significantly lower MAE, demonstrating improved learning capabilities, as evidenced by Fig~\ref{fig:train_curve}(b). The MAE obtained after approximately 500 epochs is around 9 for the training data and 29 for the testing data, reflecting a substantial improvement.

By balancing the dataset, the model gained a more representative distribution of data from the minority class, which contributed to the observed reduction in MAE.

\begin{table}[h]
    \centering
    \caption{Mean Absolute Error (MAE) for Original and Balanced Datasets}
    \begin{tabular}{llccc}
        \hline
        \textbf{Input Data} & \textbf{Dataset} & \textbf{Train MAE} & \textbf{Test MAE} & \textbf{Validation MAE} \\ \hline
        RGB \& Numeric & Original (Imbalanced) & 189.04 & 189.22 & 208.23 \\ 
        RGB \& Numeric & Balanced & 8.44 & 28.82 & 36.25 \\ \hline
    \end{tabular}
    \label{tab:mae_comparison}
\end{table}

\subsection{Frog Occurrence Classification:}
\subsubsection{Comparison of Different Input Data Sources for Frog Occurrence}
The classification accuracy and ROC AUC score for the Train, Test and Validation data are presented in Table~\ref{tab:classification_results}.  The combination of land cover (LC) and numeric data resulted in the highest accuracy for classifying locations as either presence or absence of frogs.
\begin{table}[h]
    \centering
    \caption{Classification : Performance Metrics for Different Input Source}
    \begin{tabular}{lcccc}
        \hline
        \textbf{Method} & \textbf{Train Acc. (\%)} & \textbf{Test Acc. (\%)} & \textbf{Validation Acc. (\%)} & \textbf{AUC Score} \\ \hline
        RGB \& Numeric & 79.0 & 76.1 & 75.7 & 0.82 \\ 
        LC \& Numeric & 91.0 & 90.9 & 89.1 & 0.96 \\ 
        NDVI \& Numeric & 90.9 & 90.8 & 88.4 & 0.96 \\ \hline
    \end{tabular}
    \label{tab:classification_results}
\end{table}

\subsubsection{Comparison of different pseudo-absence data generation method}
The AUC score and accuracy obtained using the two existing methods—1) Random Selection, which involves randomly assigning pseudo-absence points across the study area, and 2) Distance Criteria, which selects pseudo-absence points based solely on their distance from known presence points—are compared with those of the proposed method. For this comparison, only the LC and numeric data types were selected. The results indicate that the proposed method outperforms the other two methods. Notably, a substantial difference is observed between the proposed method and the random selection method; however, the training and testing scores for pseudo-absence data generated by the distance criteria method closely align with those of the proposed method.

This similarity can be attributed to the selection criteria employed in generating the pseudo-absence data. The proposed method incorporates both geographical extent and land cover type when selecting absence points, while the distance criteria method considers only geographical extent. Consequently, there is some inherent correlation between the two methods regarding the training and testing data, which is not reflected in the validation accuracy. This indicates that while both methods may perform similarly on training and testing datasets, the proposed method demonstrates better generalization capabilities in real-world scenarios, as evidenced by its superior validation accuracy.

\begin{table}[h]
    \centering
    \caption{Comparison of Pseudo-Absence Data Generation Methods}
    \begin{tabular}{lcccc}
        \hline
        \textbf{Method} & \textbf{Train Acc. (\%)} & \textbf{Test Acc. (\%)} & \textbf{Validation Acc. (\%)} & \textbf{AUC Score} \\ \hline
        Proposed Method & \textbf{91.0} & 90.0 & \textbf{84.8} & \textbf{0.90} \\ 
        Random Selection & 72.2 & 70.1 & 65.9 & 0.68 \\ 
        Distance Criteria & 90.4 & \textbf{90.1} & 80.1 & 0.88 \\ \hline
    \end{tabular}
    \label{tab:pseudo_absence_comparison}
\end{table}

\subsection{Frog Counting Regression}
\subsubsection{Comparison of Different Input Data Sources for Frog Counting}
\label{regression-task}

In this analysis, we trained three different models utilizing combinations of RGB \& Terraclimate data, Landcover \& Terraclimate data, and NDVI \& Terraclimate data. The results obtained on the test and validation data are presented in Table~\ref{tab:performance_comparison_reg}. From this table, it is evident that the model utilizing LC \& Numeric data outperforms the other data types, with the NDVI \& Numeric data model ranking as the second best.

The observable difference in performance can be attributed to the nature of the data. The Esri 10-meter land cover dataset categorizes the study area into ten distinct classes of land types, resembling a form of semantic segmentation. This classification enables the model to distinguish features more effectively from one point to another, facilitating easier learning. In the case of NDVI, the study area is represented by values in the range of -1 to 1, reflecting the vegetation health of the area. Similar to the Esri dataset, the model can learn features more effectively when compared to the RGB dataset.

\begin{table}[h]
    \centering
    \caption{Regression : Performance Metrics for Different Input Source}
    \begin{tabular}{lccccc}
        \hline
        \textbf{Input Data} & \textbf{Train MAE} & \textbf{Test MAE} & \textbf{Validation MAE} \\ 
        % \cline{4-5}
         % &  &  & \textbf{Sliding Window} & \textbf{Resizing} \\ 
        \hline
        RGB \& Numeric & 8.44 & 28.82  & 36.25 \\ 
        LC \& Numeric & 5.06 & 12.08  & 30.18 \\ 
        NDVI \& Numeric & 3.99 & 12.17 & 33.74 \\ 
        \hline
    \end{tabular}
    \label{tab:performance_comparison_reg}
\end{table}

\subsubsection{Weighted Average Ensemble}
Given the hardware constraints and the substantial computational overhead associated with training a single model using all modalities (RGB, NDVI, Landcover) and numerical data simultaneously, we opted for an alternative approach. Instead of combining all input modalities into one model, we employed a weighted average ensemble method to leverage the strengths of individual models trained on different data types. This approach mitigates the challenges posed by model size and training time by a more efficient integration of models trained of on different sources. The weighted average ensemble method combines the predictions of three models to produce a single optimal prediction, as detailed below.

In this approach, each model's contribution to the final prediction is weighted according to its performance. The formula for the weighted average prediction is given in Equation~\ref{ens-equ}:

\begin{equation}
y_i = \frac{(W_a \cdot P_a) + (W_b \cdot P_b) + (W_c \cdot P_c)}{W_a + W_b + W_c} \quad
\label{ens-equ}
\end{equation}

where:
- \( y_i \) is the prediction of the ensemble model.
- \( W_a, W_b, W_c \) are the weights for models \( a, b, \) and \( c \), respectively.
- \( P_a, P_b, P_c \) are the predictions from models \( a, b, \) and \( c \), respectively.

To determine the optimal weights for each model, the weights were initialized randomly in such a way that their sum equals 1 (here, $W_a=0.3$, $W_b=0.3$, and $W_c=0.4$). The weighted average predictions were calculated using Equation~\ref{ens-equ}, and the Mean Absolute Error (MAE) was computed between the true values and the predictions. An optimization algorithm was then employed to adjust the weights while satisfying the constraints that the weights sum to 1 and are bounded between 0 and 1. This optimization method iteratively updates the weights to minimize the MAE, thereby enhancing the performance of the ensemble model.

\begin{figure}
\centering
\includegraphics[width=.5\textwidth]{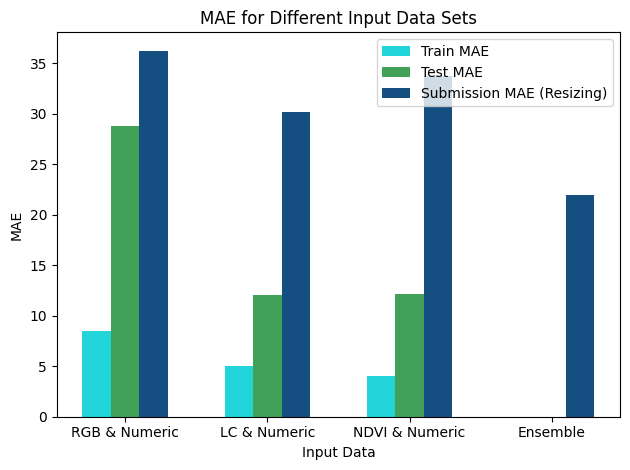}
\caption{Comparison of MAE for model trained on different input data source combinations and the ensemble model}
\label{fig:ensembling_result}
\end{figure}

The weighted average ensemble significantly improved model performance by addressing the differences observed among the individual models utilizing three different data types. The final optimized weights are as follows: RGB - 0.1, LC - 0.6, NDVI - 0.3. The model using Landcover (LC) was assigned a higher weight due to its superior performance compared to the other two models. The results obtained on the validation data are shown in Figure~\ref{fig:ensembling_result}. The figure demonstrates that the ensembled model achieved the lowest validation MAE score, reinforcing its potential for enhancing model performance in real-world applications.

\subsection{Feature Importance Assessment}
The Terraclimate dataset consists of multiple parameters, and understanding which parameters significantly influence frog presence or absence at a given location is important. Feature selection, the technique used to identify a relevant subset of features, plays a crucial role in enhancing model efficiency. Reducing the number of features not only conserves computational resources but also mitigates the risk of including irrelevant features that may lead to poorer predictive performance.

In this study, we employed Recursive Feature Elimination (RFE) to rank the importance of features. RFE iteratively fits the model, starting with all 14 parameters from the Terraclimate dataset and discarding the least important features one by one until the desired number of features is retained. A Random Forest Regressor model was utilized to rank the features based on their importance, and the target variable in this context represents frog count. The final selection comprised the top 10 features.

\begin{table}[h]
    \centering
    \caption{Feature Importance Ranking}
    \begin{tabular}{cccccccccccc}
        \hline
        \textbf{Feature} & Tmax & Tmin & Pet & Ppt & Vap & Vpd & Soil & Ws & Q & Pdsi \\ \hline
        \textbf{Importance} & 0.55 & 0.16 & 0.08 & 0.05 & 0.05 & 0.04 & 0.03 & 0.03 & 0.01 & 0.00 \\ \hline
    \end{tabular}
    \label{tab:feature_importance}
\end{table}

The results of the RFE analysis, which ranks the features according to their importance, are presented in Table~\ref{tab:feature_importance}. The analysis revealed that the majority of the parameters had negligible contributions to the target variable. To evaluate the impact of using a reduced feature set, we compared the performance of the model utilizing all features to that of a model employing only the top 6 features. The findings are summarized in Table~\ref{tab:model_performance_feature}. The results indicate that reducing the Terraclimate dataset to 6 features yielded slightly better results in terms of Mean Absolute Error (MAE) compared to using all features. Further, a slight improvement in inference time was observed when employing just the top 6 features, suggesting that feature reduction can enhance computational efficiency without significantly compromising predictive accuracy. Additionally, using fewer features allows the models to learn and map the problem more effectively, particularly in cases with limited data, such as ours.

\begin{table}[h]
    \centering
    \caption{Model Performance Comparison for Feature Importance Assessment}
    \begin{tabular}{lccccc}
        \hline
        \textbf{Input Data} & \textbf{Features} & \textbf{Train MAE} & \textbf{Test MAE} & \textbf{Validation MAE} & \textbf{Inf. Time (sec)} \\ \hline
        LC \& Numeric & 10 & 5.06 & 12.08 & 30.18 & 0.064 \\
        LC \& Numeric & 6 & 15.14 & 23.58 & 30.58 & 0.058 \\ \hline
    \end{tabular}
    \label{tab:model_performance_feature}
\end{table}

\subsection{Generalizing the Model Across Geographical Regions}
To test the proposed model’s generalization, we evaluated how well it performs in predicting frog counts in different locations, simulating real-world scenarios and assessing the model’s ability to extrapolate its knowledge to unseen environments. For this experiment, the fusion models (Model-A, Model-B, Model-C) was trained exclusively using the Australian dataset, and the trained model was then used to predict frog counts in Costa Rica. The results are presented in Table~\ref{tab:input_data_performance}, which shows that, consistent with findings in subsection~\ref{regression-task}, the combination of land cover (LC) and numeric input data produced slightly better performance. The Mean Absolute Error (MAE) achieved in this case was comparable to that of the model trained on data from all three countries, demonstrating the model's capability to generalize effectively and make accurate predictions on unseen data.

\begin{table}[h]
    \centering
    \caption{Performance Metrics for Different Input Data Types}
    \begin{tabular}{lc}
        \hline
        \textbf{Input Data}  & \textbf{Validation MAE} \\ \hline
        RGB \& Numeric  & 32.57 \\ 
        LC \& Numeric  & 32.12 \\ 
        NDVI \& Numeric  & 32.73 \\ \hline
    \end{tabular}
    \label{tab:input_data_performance}
\end{table}

\section{Implications for Ecology}
The findings of this study have significant implications for ecological research and conservation efforts, particularly in the context of species distribution modeling and frog (Anura) conservation studies. By integrating multimodal data sources, such as high-resolution imagery and environmental covariates, this study provides a robust framework for improving the accuracy and scalability of SDMs. These advancements allow researchers to better understand how environmental factors influence species distribution patterns, which is essential for assessing the impacts of climate change, habitat degradation, and other ecological stressors on the species under study.

For frogs in particular, which are highly sensitive bio-indicators of ecosystem health~\cite{iman2020MarshF, kurnianto2024assessing}, the developed SDM offers a valuable tool for monitoring population trends and identifying critical habitats. The ability to accurately classify areas of presence and absence, combined with improved frog counting predictions, enables more effective conservation planning. This is particularly important given the global decline in amphibian populations is suspected due to pollution, climate change, and habitat loss~\cite{Collins2010-gz, frog_decline1}. The pseudo-absence data imputation method proposed in this study further enhances the reliability of ecological predictions, addressing a key limitation in traditional SDM approaches. 

Moreover, the methodology's generalization capabilities, demonstrated by its strong performance across diverse geographic regions, add an potential for application in broader ecological contexts. By optimizing input features and employing data balancing techniques, the proposed approach ensures that limited or incomplete datasets can still produce meaningful results. 

Finally, the proposed method enables the detection of biodiversity hotspots regions of rich biodiversity that warrant extra attention due to their ecological significance. Identifying these hotspots can aid in prioritizing conservation efforts and focusing resources on areas that are critical for maintaining ecosystem health and preserving species diversity. This is particularly relevant for conservation policy-makers, who require precise and scalable tools to prioritize areas for intervention and to develop strategies for preserving biodiversity.

\section{Conclusion}
This research aimed to develop a robust multimodal Species Distribution Model (SDM) for Anura using diverse data sources. Traditional SDMs often rely on single-modality inputs and face significant limitations, such as incomplete data and imbalance across classes. By integrating multimodal learning techniques, this study demonstrates a scalable approach for addressing these challenges. The proposed SDM effectively handles both frog counting and presence/absence classification tasks, showcasing its ability to generalize across unseen data.

The use of a late fusion architecture integrating image and tabular data was key to achieving superior results, with the model attaining an AUC score of 0.90 and an accuracy of 84.9\% in presence/absence classification. Additionally, the proposed pseudo-absence data imputation method enhanced the reliability of absence data, while adaptive data balancing techniques ensured diversity and adequate representation across classes.

These outcomes not only demonstrate the potential of multimodal learning for ecological predictions but also highlight the practical importance of addressing key data challenges. By leveraging advanced deep learning techniques, this study contributes to the development of precise and scalable SDMs, offering valuable tools for conservation policy-makers and researchers to better understand and protect fragile ecosystems.

\subsection{Future Work}

\subsubsection*{Pseudo-Absence Data Generation} The method used for generating pseudo-absence data can be refined to include additional variables. For instance, the current method considers only the geographical extent and land cover patches when selecting pseudo-absence points. Parameters such as temperature and precipitation, which have proven to be important factors influencing frog habitats, should also be considered.

\subsubsection*{Including Historical Data for Better Representation} The frog presence dataset provided includes data from the years 2017 to 2019, and only the covariates from these periods are considered when building the SDM. However, incorporating climatic data from previous years could generate additional data and better represent higher frog counts. This approach could be explored in the future to determine whether it leads to improved predictions.

\subsubsection*{SDM Based on Graph Neural Networks (GNN)} A similar approach to the GNN method explained in Section3.3 for weather forecasting could be applied to building an SDM. The dataset has already been prepared using the method described in Section4.4, but experiments could not be conducted within the available timeframe. This direction of research could yield better results in the future.

%%%%%%%%%%%%%%%%%%%%%%%%%%%%%%%%%%%%%%%%%%%%%%
%%                                          %%
%% Backmatter begins here                   %%
%%                                          %%
%%%%%%%%%%%%%%%%%%%%%%%%%%%%%%%%%%%%%%%%%%%%%%

\begin{backmatter}

\section*{Abbreviation}
SDM: Species Distribution Model; GNN: Graph Neural Networks; CNN: Convolutional Neural Network; ResNet: Residual Network; NDVI: Normalized Difference Vegetation Index; NDWI: Normalized Difference Water Index;  ROC: Receiver Operating Characteristic; RFE: Recursive Feature Elimination; AUC: Area Unver Curve; DNN: Deep Neural Network;.

\section*{Acknowledgments}
 Andreas Kamilaris, and Chirag Padubidri have received funding from the European Union's Horizon 2020 Research and Innovation Programme under grant agreement No. 739578 complemented by the Government of the Republic of Cyprus through the Directorate General for European Programmes, Coordination and Development.
 
\section*{Funding}

\section*{Availability of data and materials}

\section*{Ethics approval and consent to participate}

\section*{Consent for publication}
Not applicable.
\section*{Competing interests}
All authors declare that they have no competing interests.
\section*{Authors' contributions}
Pranesh Velmurugan carried out the experiment under the guidance of Chirag Padubidri. Chirag Padubidri wrote the manuscript with support from Andreas Kamilaris, and Pranesh Velmurugan. Andreas Kamilaris supervised the project.

%%  After submission of the .TEX file,                     %%
%%  you will be prompted to submit your .BBL file.         %%

% if your bibliography is in bibtex format, use those commands:
\bibliographystyle{bmc-mathphys} % Style BST file (bmc-mathphys, vancouver, spbasic).
\bibliography{bmc_article}      % Bibliography file (usually '*.bib' )
% for author-year bibliography (bmc-mathphys or spbasic)
% a) write to bib file (bmc-mathphys only)
% @settings{label, options="nameyear"}

% or include bibliography directly:
% \begin{thebibliography}
% % \bibitem{b1}
% \end{thebibliography}

\end{backmatter}
\end{document}